\documentclass[lettersize,journal]{IEEEtran}
\usepackage{amsmath,amsfonts}
\usepackage{algorithmic}
\usepackage{algorithm}
\usepackage{array}
\usepackage[caption=false,font=normalsize,labelfont=sf,textfont=sf]{subfig}
\usepackage{textcomp}
\usepackage{stfloats}
\usepackage{url}
\usepackage{verbatim}
\usepackage{graphicx}
\usepackage{xcolor}

\usepackage{cite}
\usepackage{amsmath,amssymb,amsfonts}
\usepackage{algorithmic}
\usepackage{graphicx}
\usepackage{textcomp}
\usepackage{xspace}
\usepackage{multirow}
\newcommand{\ie}{\emph{i.e.,}\xspace}
\newcommand{\eg}{\emph{e.g.,}\xspace}
\newcommand{\eat}[1]{}

\newcommand{\model}{{VARS}\xspace}
\newcommand\etal{\emph{et al.}}

\colorlet{blue}{black}
\begin{document}

\title{Versatile and Risk-Sensitive Cardiac Diagnosis via Graph-Based ECG Signal Representation}

\author{Yue Wang,  Yuyang Xu, Renjun Hu, Fanqi Shen, Hanyun Jiang, Jun Wang, Jintai Chen, Danny Z. Chen \IEEEmembership{Fellow, IEEE}, Jian Wu, \IEEEmembership{Member, IEEE}, and Haochao Ying, \IEEEmembership{Member, IEEE}
\thanks{
This research was supported by the National Natural Science Foundation of China under Grant No.~62476246, the Zhejiang Provincial Natural Science Foundation of China under Grant No.~LY23F020019, and the Opening Foundation of the State Key Laboratory of Transvascular Implantation Devices under Grant No.~SKLTID2024003.
 (Corresponding Authors: Haochao Ying and Jian Wu.)}
\thanks{
Yue Wang and Yuyang Xu are with the College of Computer Science and Technology, Zhejiang University, Hangzhou 310012, China. They are also with the State Key Laboratory of Transvascular Implantation Devices of the Second Affiliated Hospital, Zhejiang University School of Medicine, Hangzhou 310009, China, and Zhejiang Key Laboratory of Medical Imaging Artificial Intelligence, Hangzhou 310058, China. (E-mail: ywang2022@zju.edu.cn, xuyuyang@zju.edu.cn).}
\thanks{Renjun Hu is with the Alibaba Group, Hangzhou 310023, China. }
\thanks{Fanqi Shen and Hanyun Jiang are with the College of Computer Science and Technology, Zhejiang University, Hangzhou 310012, China.}
\thanks{Jun Wang is with the School of Computer and Computational Science, Hangzhou City University, Hangzhou 310015, China. }
\thanks{Jintai Chen is with the Information Hub, Hong Kong University of Science and Technology (Guangzhou), Guangzhou, 511458, China.}
\thanks{Danny Z. Chen is with the Department of Computer Science and Engineering, University of Notre Dame, Notre Dame, IN 46556, USA.}
\thanks{Jian Wu is with the State Key Laboratory of Transvascular Implantation Devices of the Second Affiliated Hospital and School of Public Health, Zhejiang University School of Medicine, Hangzhou 310009, China. He is also with Zhejiang Key Laboratory of Medical Imaging Artificial Intelligence, Hangzhou 310058, China.}
\thanks{Haochao Ying is with the School of Public Health and Second Affiliated Hospital, Zhejiang University School of Medicine, Hangzhou 310058, China. (E-mail: haochaoying@zju.edu.cn)}
}


\markboth{Journal of \LaTeX\ Class Files,~Vol.~14, No.~8, August~2021}%
{Shell \MakeLowercase{\textit{et al.}}: A Sample Article Using IEEEtran.cls for IEEE Journals}


\maketitle
\begin{abstract}
Despite the rapid advancements of electrocardiogram (ECG) signal diagnosis and analysis methods through deep learning, two major hurdles still limit their clinical adoption: the lack of versatility in processing ECG signals with diverse configurations, and the inadequate detection of risk signals due to sample imbalances. 
Addressing these challenges, we introduce \underline{V}ers\underline{A}tile and \underline{R}isk-\underline{S}ensitive cardiac diagnosis (\model), an innovative approach that employs a graph-based representation to uniformly model heterogeneous ECG signals. \model stands out by transforming ECG signals into versatile graph structures that capture critical diagnostic features, irrespective of signal diversity in the lead count, sampling frequency, and duration. This graph-centric formulation also enhances diagnostic sensitivity, enabling precise localization and identification of abnormal ECG patterns that often elude standard analysis methods.
To facilitate representation transformation, our approach integrates denoising reconstruction with contrastive learning to preserve raw ECG information while highlighting pathognomonic patterns. 
We rigorously evaluate the efficacy of \model on three distinct ECG datasets, encompassing a range of structural variations. The results demonstrate that \model not only consistently surpasses existing state-of-the-art models across all these datasets but also exhibits substantial improvement in identifying risk signals. Additionally, \model offers interpretability by pinpointing the exact waveforms that lead to specific model outputs, thereby assisting clinicians in making informed decisions. 
These findings suggest that our \model will likely emerge as an invaluable tool for comprehensive cardiac health assessment.
\end{abstract}

\begin{IEEEkeywords}
ECG Representation, Cardiac Diagnosis, Denoising Reconstruction, Contrastive Learning
\end{IEEEkeywords}

\section{Introduction}

Electrocardiogram (ECG) stands as the most accessible and cost-effective technique for detecting a wide range of cardiac anomalies, such as atrial fibrillation, myocardial ischemia, and hypokalemia, among others~\cite{siontis2021artificial,fernandez2019artificial,gadaleta2023prediction,kalmady2024development,lee2024artificial,wang2023ecggan}. This diagnostic method involves the attachment of electrodes at specified locations on the body skin near the heart to monitor the electrical activity of the heart muscle. 
In practice, ECG devices typically utilize between 2 to 12 leads, acquiring data at sampling frequencies ranging from 250 to 1000 Hz for varying durations.
Clinicians then rely on these data to diagnose conditions such as arrhythmias and structural cardiac abnormalities by scrutinizing crucial yet frequently subtle aspects like heart rhythm, waveform morphology, and timing intervals between beats. Not surprisingly, manual diagnosis is time-consuming and requires high-level expertise. Consequently, the development of model-based approaches for automated cardiac diagnosis has recently emerged as a significant area of research.

However, the heterogeneity of ECG data, accompanied by patient-specific variables and a diverse array of monitoring equipment, introduces significant challenges to diagnosis. A spectrum of sophisticated models, especially those based on Deep Neural Networks (DNNs), has been proposed.
Despite the effectiveness of DNNs for ECG diagnostics~\cite{mei2019detecting,9246692,liu2021deep,al2023machine,denysyuk2023algorithms,zhang2023feature}, their applications remain constrained to analysis within specific data types, owing to employing pre-defined feature extraction modules and classification labels. The fundamental differences in the number of leads, sampling frequencies, and durations of ECG recordings have systematically challenged the versatility of the representational capabilities of traditional DNN models~\cite{hong2020opportunities}. It turns out that, while these models perform admirably on certain datasets, their ability to consistently recognize and differentiate intricate patterns across diverse ECG signals remains questionable~\cite{zhou2019k}. 
Additionally, their performance is compromised in the face of class imbalance within ECG datasets, indicating a notable deficiency in detecting risk signals~\cite{rath2021heart}. The over-reliance on labels further undermines the model performance and their capacity to generalize, impeding the learning of effective ECG signal representations. In summary, current ECG signal representation and processing approaches exhibit two critical limitations: the lack of versatility in processing ECG signals with diverse configurations, and the inadequate detection of risk signals due to sample imbalances.

\begin{figure}[t!]
    \centering
    \includegraphics[width=0.96\linewidth]{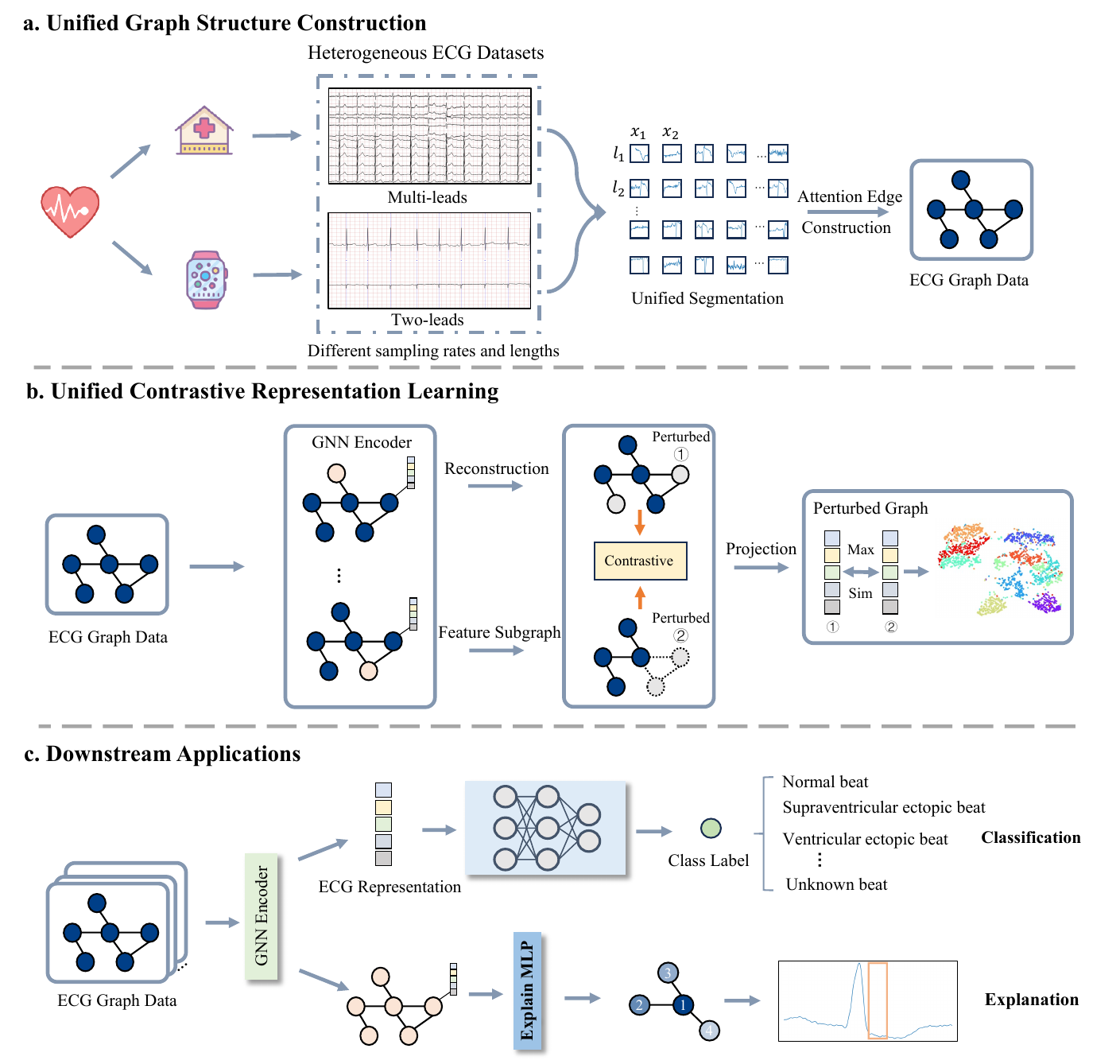}
    \caption{\textbf{Illustrating our \model approach.} The approach involves three parts: unified structure construction, representation learning, and downstream tasks. \textbf{a.} Heterogeneous ECG datasets, including multi-lead and two-lead configurations with varied sampling rates and lengths, are transformed into a unified signal structure to ensure consistency. \textbf{b.} A contrastive learning method with a Graph Neural Network (GNN) encoder learns a unified ECG representation, with feature subgraphs and perturbed graphs, enhancing robustness. \textbf{c.} These graph structures support downstream tasks like classification and interpretation, categorizing signals into normal beats, supraventricular ectopic beats, ventricular ectopic beats, and other types. }
    \label{illustrating}
\end{figure}

To overcome these challenges, we propose an innovative approach, \underline{V}ers\underline{A}tile and \underline{R}isk-\underline{S}ensitive cardiac diagnosis (\model), which leverages a graph-based representation to handle the heterogeneity of ECG signals for versatile and risk-sensitive cardiac health assessment.
We develop a new method for transforming ECG signals into a universal graph structure. To achieve the goal of unified data abstraction, we segment an ECG sequence with a specific time interval, and each time interval is taken as a graph node. 
Correspondingly, different leads are converted into different sets of nodes.
This approach aligns well with cardiologists' clinical knowledge viewing the features of a time interval (PR interval) as a pattern unit to assess diseases. 
We then calculate the distances between the nodes to compose the ECG graphs so that semantic correlations between raw signal sequences are preserved.
This graph-based abstraction has the potential to capture key diagnostic features while remaining unaffected by variations in lead count, sampling frequency, and signal duration.

ECG data often present the `lost-in-the-middle' quandary, where critical diagnostic decisions hinge upon a minority of waveforms within the full recording. These small yet significant patterns are known to be difficult to detect, especially in the situations of class imbalances. As illustrated in Figure~\ref{illustrating}, our graph-centric perspective 
aims to increase diagnostic sensitivity, enabling precise localization and identification of those elusive abnormal ECG patterns that conventional methods struggle to discern.
Specifically, \model  combines denoising reconstruction~\cite{hou2022graphmae} and contrastive learning~\cite{wu2019session} for this purpose.
Denoising reconstruction can filter out irrelevant or misleading noise from raw signals without damaging the integrity of the original data, hence preserving the original ECG information while enhancing representational effectiveness. This step is crucial for maintaining the signals' purity and usability. 
Additionally, contrastive learning plays an essential role in enhancing the model's ability to differentiate among diagnostic patterns. By comparing different signal patterns and amplifying the defining features of each, this technique facilitates a more detailed and nuanced understanding of the data.
However, classic contrastive learning methods cannot effectively capture subtle key information from the entire ECG graphs. To address this, we implement subgraph extraction and representation for contrastive learning. In this way, we enhance discrimination among signals and widen the distance between samples of different categories in the feature space, thereby improving the model's representation ability for classification.

We conduct extensive experiments on three distinct ECG datasets (\ie MITBIH, PTB-XL, and ST-T) that encompass a range of structural variations to evaluate the efficacy of \model. 
We show that \model consistently surpasses 12 baselines across these datasets in all the tested metrics. Moreover, \model also exhibits more substantial improvement in identifying risk signals. For instance, 
in the anomaly categories, \model outperforms the baselines by a factor of ten compared to its improvement in the overall classification. Additionally, \model enhances the F1 score by $\sim$6.4\% across the three datasets for the risk categories.
We present an ablation study to empirically justify the technical designs behind \model. Finally, our case study and visual analysis results further confirm the advantages of \model in interpretability (by pinpointing the exact waveforms that lead to specific model outputs) and representation (with better categorical separation).

The main contributions of our work are as follows:
\begin{itemize}
    \item We study versatile and risk-sensitive cardiac diagnosis and propose a novel transformation from ECG signals to graphs, which is among the first in the literature.  
    
    \item We develop the new \model approach that integrates denoising reconstruction and contrastive learning to facilitate the representation transformation and signal analysis.
    
    \item We conduct extensive experiments to demonstrate the multi-dimensional superiority of \model on versatility for signal diversity, risk-sensitivity, and interpretability.
\end{itemize}

\section{Related Work}

\subsection{Graph Structure Construction}
When applying Graph Neural Networks (GNNs) to data that do not directly provide a graph structure, it is essential to construct a graph representation to enable effective GNN processing. Typically, there are two main types of approaches for constructing graphs.
One type involves calculating the similarity between nodes based on their features within the dataset, while the other type constructs a graph by leveraging the explicit physical associations among the data elements.
In the first type, Vision GNN~\cite{han2022vision} divided an image into multiple patches, converted each patch into a one-dimensional vector to form a node representation of the image, and computed $k$-nearest neighbors of each node to form edges between the nodes. Time2Graph~\cite{cheng2020time2graph} determined the weights of edges between nodes by calculating the distribution probabilities between nodes to detect anomalies in time series data. BrainNet~\cite{chen2022brainnet} calculated cosine similarity between nodes and set a threshold to establish edges between nodes. GraphSleepNet\cite{jia2020graphsleepnet} constructed the sleep stage network using an adjacency matrix learned from EEG signal sequences, and the sleep feature matrix was filled by differential entropy features at each frequency band.
In the second type, GDN~\cite{deng2021graph} computed the normalized dot product between the embedding vectors of sensors to learn the relationships between the sensors according to the realistic positions of the sensors. MST-GAT \cite{ding2023mst} used time series embedding to construct a flexible graph structure and computed cosine similarity to build an adjacency matrix to adequately represent the time series' inherent properties. 
In this paper, different from the prior work, we investigate the capability of graph construction to tackle the problem of handling variations in ECG signals effectively. Our graph structure preserves specific time-interval information and captures sequential correlations within the signals, thereby enabling its potential application in ECG disease diagnosis.

\subsection{ECG  Representation Learning}

Many artificial intelligence-based methods have been developed to enhance the representations of ECG signals, improving both the accuracy and efficiency of detecting cardiac abnormalities~\cite{biton2023generalizable,yehuda2023self,lee2022efficient}. ECG representation learning methods primarily fall into two categories: supervised learning based on class labels and unsupervised learning. In supervised representation learning, models focus on capturing the spatial and temporal features of ECG data by extracting patterns associated with class labels. For instance, RCNNs~\cite{le2021multi,wang2023single,islam2023hardc,mason2024ai} developed a hybrid model combining CNN and RNN modules to learn suche features, enabling accurate classification of one-lead fixed-length ECG beats. Similarly, 1D-CNN~\cite{yildirim2018arrhythmia,goodfellow2018towards,niu2019inter,kolk2024deep} was adapted for arrhythmia classification using long-duration ECG signals (e.g., ten seconds). An improved Residual Network~\cite{ribeiro2020automatic,allam2020spec} was applied to standard short-duration 12-lead ECG data, outperforming cardiology residents in detecting six types of abnormalities. \eat{MINA~\cite{hong2019mina} achieved ECG representation by detecting rhythm and frequency abnormalities. }
Zubair \etal~\cite{zubair2023deep} employed a re-sampling strategy with a translation loss function to enhance focus on relevant information for arrhythmic beat detection.

However, an overemphasis on class label information can fall short of providing a comprehensive characterization of ECG signals, leading to insufficient representations. In contrast, unsupervised representation learning seeks to address this issue by capturing the intrinsic nature of ECG signals. Nakamoto \etal~\cite{nakamoto2022self} and Rabbani \etal~\cite{rabbani2022contrastive} employed contrastive learning to bring ECG signals closer to their augmented samples in the feature space while pushing them further away from the other ECG signals and their augmented samples. Grabowski \etal~\cite{grabowski2022classification}, Rodriguez \etal~\cite{vazquez2022transformer}, and Gedon \etal~\cite{gedon2021first} applied random masking to ECG signals and trained models to predict the masked portions, extracting features for downstream tasks.
\eat{SSD~\cite{raj2018sparse} utilized sparse decomposition through an overcomplete Gabor dictionary for efficient feature extraction.} 
Chen \etal~\cite{chen2022me} and Hu \etal~\cite{hu2024personalized} proposed methods to integrate disease information into the generative process, yielding ECG signal representations enriched with disease-specific features. While previous methods often captured fundamental signal patterns, they still struggled with detecting risk signals in imbalanced ECG data, where crucial diagnostics may rely on a few critical waveforms.  This is because standard augmentations in unsupervised learning, such as random dropping and noise addition, often miss small but significant patterns, making them harder to detect risk signals.


\section{Method}
\begin{figure*}[t!]
    \centering
    \includegraphics[width=0.99\textwidth]{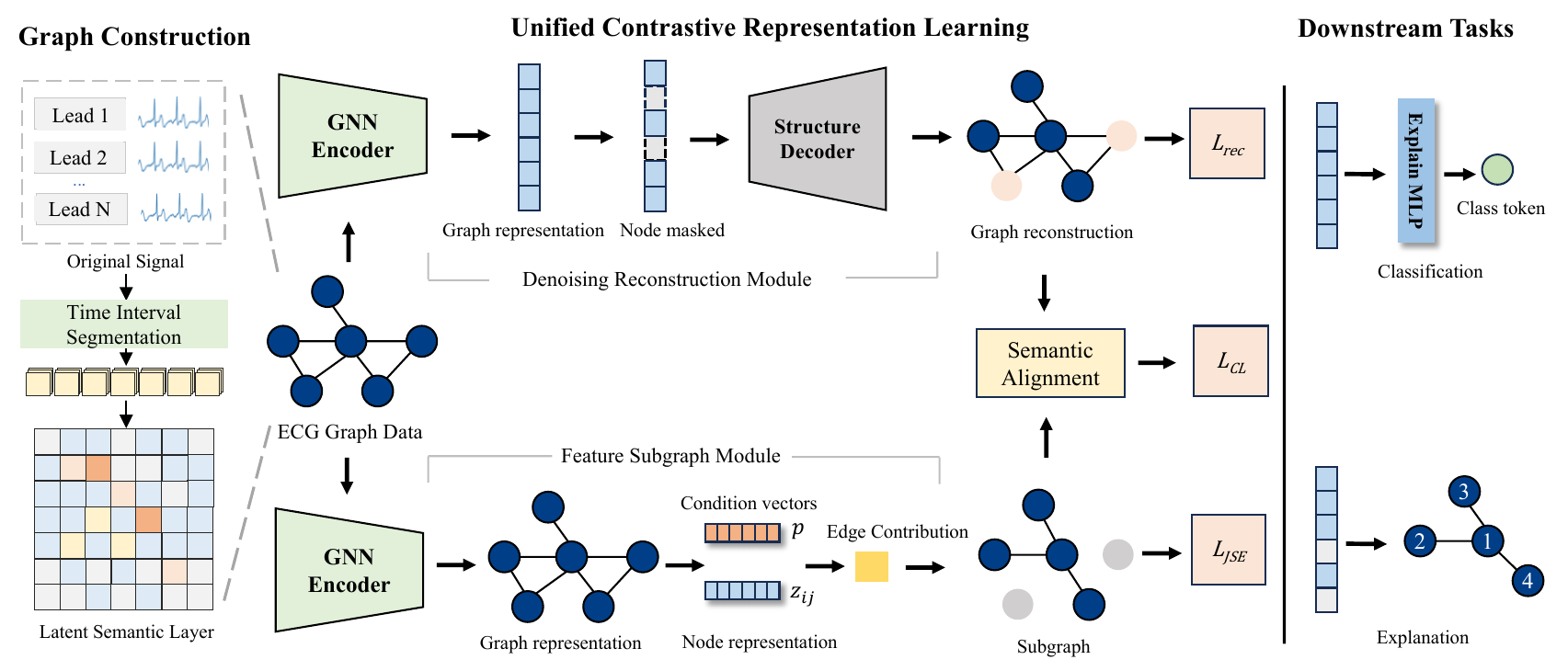}
    \caption{\textbf{An overview of our \model framework.} \model consists of three main parts. The graph construction process transforms ECG data into graph data, using attention mechanisms to capture semantic relationships between graph nodes. Feature subgraph contrastive learning is employed to generate a unified ECG representation. Downstream tasks utilize the trained GNN encoder for classification and interpretation of graph data. }
    \label{fig:Structure}
\end{figure*}
\subsection{Problem Definition}

We first describe the notation and definition of the problem in this study. Suppose an ECG dataset $S = \left[S_1, S_2, \ldots, S_r\right]$ is given (\( S \) contains \( r \) samples). Each sample \( S_i = [s_1, s_2, \) $\ldots, s_n] \in \mathbb{R}^{n \times t}$ is recorded from \( n \) leads, with \( t \) timestamps per lead. We divide each lead of $S_i$ into a sequence of time intervals, \( s_l = \left[x_1, x_2, \ldots, x_{\frac{t}{m}}\right] \), i.e., $s_l$ is represented as \( s_l \in \mathbb{R}^{\frac{t}{m} \times m} \), with \( m \) denoting the preset length of the time interval.

\textbf{Arrhythmia Classification.} Given the ECG dataset ${S}$ and the corresponding label set ${Y}=\left[{y}_{{1}}, {y}_{2}, \ldots, {y}_r\right]$, where $y_v \in [{0,1}]$ is the value of the ${v}$-th label. The objective of arrhythmia classification is to predict the labels $\hat{Y}$ of an ECG test set ${{S}_{test}}$:
\begin{equation}
    \centering
    \hat{Y}=M\left( {Y},  R({S}),R({S_{test}})\right),
\end{equation}
where 
${R}$ is a representation learning model used to extract features from ECG signals, and ${M}$ is a linear classifier.


\subsection{General Description}


To develop a unified representation of ECG data and 
achieve versatile and risk-sensitive cardiac diagnosis, we propose a novel unified ECG representation approach, \model.
Figure~\ref{fig:Structure} illustrates the structure of the \model model, mainly comprising three primary modules: the graph construction module, the denoising reconstruction module, and the feature subgraph contrastive learning module.
First, to address the issue of non-uniform ECG data formats, \model transforms the heterogeneous ECG data with different leads into graph data. In the graph construction module, we use a specific time interval to define a graph node, and the signal patterns between nodes are computed with a self-attention mechanism.
Second, to enhance risk-sensitive cardiac diagnosis, we propose a unified representation contrastive learning method. 
Specifically, we employ a GNN decoder to reconstruct masked graphs, aiming to preserve the signals' purity and usability while simultaneously filtering out irrelevant noise.  On the other hand, by leveraging the feature subgraph contrastive learning module, we extract subgraphs to obtain sample pairs to enhance discrimination among signals 
in contrastive learning.
Finally, the trained GNN encoder is employed for downstream tasks such as classification and interpretation of graph data.


\subsection{Graph Construction}

To address the challenge of processing ECG signals with diverse configurations, we develop a method to convert ECG signals into a universal graph structure. By establishing nodes and edges, the graph structures allow us to preserve essential temporal interval details and accurately capture sequential relationships within the data. Moreover, considering the potential complex dependencies and diagnostic patterns among different time intervals in ECG signals, we introduce a self-attention mechanism to automatically learn these patterns. The self-attention mechanism enables the model to flexibly capture long-range dependencies in sequential data, hence more accurately uncovering significant patterns within ECG signals. In this method, we compute attention scores to enable the mechanism to focus on the most relevant aspects of the signals. Specifically, we apply multi-head self-attention to the set of nodes representing the time intervals. Each attention head computes its own attention score matrix, capturing different aspects of the inter-node relationships. For each attention head $e$, we compute the attention score matrix $\mathrm{W_e}$ using the query \(\mathrm{Q}_e\) and key \(\mathrm{K}_e\) matrices, as follows:
\begin{equation} \mathrm{W_e}=\operatorname{Sigmoid}\left(\frac{\mathrm{Q_eK_e}^{\mathrm{T}}}{\sqrt{\mathrm{d}_{\mathrm{k}}}}\right), 
\end{equation}
where 
$d_k$ is the hidden dimension size of $Q_e$ and $K_e$.
Finally, we take the mean of the attention score matrices, $\overline{W}$, which is used to compute our adjacency matrix $A$.

To construct the adjacency matrix \(A\) , we apply a threshold \(\Theta\) to \(\overline{W}\), converting scores below the threshold to zeros:
\begin{equation}
A_{ij} = \begin{cases} 
\overline{W}_{ij}, & \text{if } \overline{W}_{ij} \geq \Theta, \\
0, & \text{otherwise},
\end{cases}
\end{equation}
where \(A_{ij}\) is the element of the adjacency matrix \(A\) for the edge between nodes \(i\) and \(j\). We use the threshold \(\Theta\) to filter out weaker connections, hence ensuring that only significant relationships are represented in the graph and yielding a sparser graph that retains only edges indicative of strong semantic correlations between nodes. This process not only simplifies the graph but also enhances the focus on semantic information.

We construct a graph $G=(\mathcal{V}, A, X)$, where$\mathcal{V}$ denotes the nodes corresponding to the time intervals, the adjacency matrix \(A \in \{0,1\}^{N \times N}\) represents edges between the nodes, 
\( X \in R^{N \times d}\) is the feature matrix of the nodes, and $N=|\mathcal{V}|$. 

\subsection{Unified Contrastive Representation Learning}
In traditional contrastive learning, graph enhancement often involves randomly masking nodes, which can potentially alter the original attributes and may lead to the creation of anomalous sample pairs. To address this issue, we propose Unified Contrastive Representation Learning, which preserves the original attributes of the data by removing irrelevant noise from normal nodes while highlighting risk signals (i.e., removing noise while maintaining the original graph structure). Initially, we perform masking operations on ECG graph data, followed by weak enhancement through the reconstruction of ECG data using the GNN decoder. Subsequently, we employ the feature subgraph contrastive learning module to strongly enhance the data, obtaining enhanced subgraphs while ignoring irrelevant noise, reinforcing signal differentiation, and amplifying the separation between samples of different categories in the feature space, thus ensuring the acquisition of more representative sample pairs.

\subsubsection*{Denoising Reconstruction Module}
To preserve the original ECG information and mitigate the impact of noise during contrastive learning, we employ a GNN encoder and GNN decoder to reconstruct the structure of the ECG graph. By reconstructing the masked nodes, we produce a perturbed graph \( G' \) that retains the semantic information of the original signals. In the GNN encoder stage, consider the graph $G=(\mathcal{V}, A, X)$
with $N=|\mathcal{V}|$.
We seek to reconstruct the features and structure, as:
\begin{equation}
H=f_E(\mathcal{V},A, X),
\end{equation}
\begin{equation}
G^{\prime}=f_D(A, H),
\end{equation}
where \( f_E \) is a graph encoder function that encodes the features of the nodes to obtain the hidden feature \( H \), and \( f_D \) is a graph decoder function that reconstructs from \( H \) a graph \( G' \), so that \( G' \) well approximates the structure and features of the original graph \( G \), ensuring essential information being preserved while reducing noise introduced during contrastive learning.

\subsubsection*{GNN-based Structure Decoder}


In general, a decoder reconstructs the original input from the latent representation generated by the corresponding encoder. In traditional natural language processing (NLP), decoders aim to recover masked words with rich semantic information, making conventional decoders like Multi-Layer Perceptrons (MLPs) highly effective. However, in ECG analysis, the subtle differences in shape and relationships between various cardiac beat structures often lead to nearly identical features when using low-level decoders. Therefore, \model employs a Graph Isomorphism Network (GIN)\cite{xu2018powerful} as its decoder to discriminate more effectively among different beat structures. The decoder correlates the nodes' and neighbors' features to reconstruct nodes, rather than solely relying on node-specific features, thus enabling the encoder to learn higher-level latent semantic representations. We define \(\tilde{\mathcal{V}} \subset \mathcal{V}\) as the subset of nodes selected for masking during training. We replace the features of these masked nodes with a learnable vector denoted as \(h_{[M]}\). The masking method is as follows:
\begin{equation}
\tilde{h}_i = \left\{
\begin{array}{ll}
h_{[M]}, & \text{if } v_i \in \tilde{\mathcal{V}}, \\
h_i, & \text{if } v_i \notin \tilde{\mathcal{V}},
\end{array}
\right.
\end{equation}
\noindent where $\tilde{h}_i$ is the hidden feature of node $v_i$ after the hidden coding masking, \( h_i \) represents the feature vector of node  \( v_i \) before masking is applied, and $h_{[M]}$  
represents the hidden coding of the nodes in the masked subset $\tilde{\mathcal{V}}$.

\subsubsection*{Feature Subgraph Module}
To address the limitations of classic contrastive learning in capturing subtle key information in ECG data, we use the embedded explainable module~\cite{xie2022task} as part of our feature subgraph module to enhance paired samples for contrastive learning. The feature subgraph module can discard unimportant edges while retaining important node information, preventing the loss of node attributes during the graph enhancement process. Simultaneously, the feature subgraph module can also interpret the classification results in downstream tasks. The formulas used by the feature subgraph module are:
\begin{equation}
G_{\text {sub }}=\left(\mathcal{V}, E_{\text {sub }}\right)=\mathcal{T}_\theta({p},G),
\end{equation}
\begin{equation}
E_{s u b}=\left\{\left(v_i, v_j\right) \ | \ \left(v_i, v_j\right) \in E \ {\rm and} \ w_{i j} \geq \delta \right\},
\end{equation}
\noindent where $\mathcal{V}$ is the node set of $G$, the model $\mathcal{T}_\theta$ determines whether to keep an edge by calculating the importance of the edge, \(\theta\) denotes the parameters of the model $\mathcal{T}_\theta$, and $p$ is a conditional vector following the Laplace distribution which is used to highlight specific node features. The edge set $E_{sub}$ is determined by applying a thresholding operation with a threshold $\delta$ to the edge contributions of the edge set $E$ of $G$. 
For an edge $(v_i,v_j) \in E$, the edge contribution $w_{ij}$ is calculated based on the features 
of nodes $v_i$ and $v_j$ in $G$.



When conducting subgraph extraction, the feature subgraph module takes the node representations $z_i$ and $z_j$ of nodes $v_i$ and $v_j$ and the conditional vector $p$ as input, and calculates the score $w_{i j}$ for an edge $(v_i,v_j) \in E$, as:
\begin{equation}
w_{i j}=\operatorname{MLP}\left(\left[\boldsymbol{z}_i ; \boldsymbol{z}_j\right] \otimes \sigma\left(f_g({P})\right)\right),
\end{equation}
\noindent where $[ \cdot; \cdot ]$ denotes concatenation along the feature dimension, $\sigma$ is the Sigmoid function, $f_g$ is a linear projection used to match the dimensionality of the data,
and $P$ is the concatenation of the result of the classification prediction and the gradient in the inference stage of downstream tasks, which provides task-specific information.  This process enables the feature subgraph module to learn edge scores related to both the graph structure and the specific requirements of the task, making the subgraph representation of ECG data more effective for downstream tasks such as classification and interpretation.

\subsection{Loss Functions}

Finally, to achieve the goal of discriminating between risk and non-risk patterns within diverse and complex ECG signals, we introduce a novel combination of three distinct loss functions, which are tailored to the denoising reconstruction, feature subgraph module, and unified contrastive representation.

In the denoising node reconstruction module, although ECG signals share similar structural types, their overall morphologies may vary significantly.  This variability poses challenges to traditional reconstruction methods, in which Mean Squared Error (MSE) often struggles to capture the nuances of such diverse patterns due to its focus on minimizing differences of fixed values.  To better capture these broader ECG patterns, we instead employ cosine error, which emphasizes the similarity in shape and overall morphology, rather than relying solely on precise value matching.

To better align the cosine similarity with the characteristics of the ECG signal reconstruction task, we scale the cosine error using a parameter $\gamma \geq 1$. Since predictions with high confidence typically result in cosine errors less than 1,  this scaling reduces the weights of such high-confidence predictions in the overall loss, thus shifting the focus toward lower-confidence predictions. For a node $v_i \in \tilde{\mathcal{V}}$, given the original node feature $x_i$ and the reconstructed output $\tilde{x_i}$ in $G^{\prime}=f_D(A, {H})$, the scaled cosine error is computed as:
\begin{equation}
    {L_{r e c}}=\frac{1}{|\tilde{\mathcal{V}}|} \sum_{v_i \in \tilde{\mathcal{V}}}\left(1-\frac{x_i \tilde{x_i}^T}{\left|x_i\right| \cdot\left|\tilde{x_i}\right|}\right)^\gamma, \gamma \geq 1,
\end{equation}
where the scale factor $\gamma$ is a tunable hyperparameter that can be optimized for ECG datasets of varying sizes.
This approach applies a global scaling to the reconstruction error, adjusting the sensitivity of the loss function and improving the model's generalization across the entire dataset.

When calculating the subgraph loss in the feature subgraph module, we use the Jensen-Shannon Estimator (JSE) to compute the loss for subgraph generation. The corresponding formula is:
\begin{align}
 {L_{J S E}}=&\min _\theta \{\frac{1}{L} \sum_{i=1}^L \log \left[\sigma\left(\left({p} \otimes \boldsymbol{H}_i\right)\left({p} \otimes \boldsymbol{H}_{i, \theta}\right)^T\right)\right] \\ \notag
 &+\frac{1}{L^2-L} \sum_{i \neq j} \log \left[1-\sigma\left(\left({p} \otimes \boldsymbol{H}_i\right)\left({p} \otimes \boldsymbol{H}_{j, \theta}\right)^T\right)\right]\},
\end{align}
\noindent where $L$ is the number of samples in a batch, $\sigma$ is the Sigmoid function, and $H_i$ and $H_{i,\theta}$ are the embeddings of the original graph $G_i$ (which is the \( i \)-th graph in a batch) and its subgraph $\mathcal{T}_\theta(p,G_i)$ respectively.

After the ECG graph data are processed by the denoising reconstruction module and the feature subgraph module, two augmented graphs are generated that retain the pathognomonic patterns of the ECG signals. These two augmented graphs are paired as positive samples, while contrastive learning is enforced by maximizing their distinction from negative sample pairs. To ensure consistency, we use the normalized temperature-scaled cross-entropy loss (NT-Xent) \cite{chen2020simple} to calculate the contrastive loss, which is given as:
\begin{equation}
L_{CL} = -\frac{1}{L} \sum_{i=1}^{L} \log \frac{\exp \left(\operatorname{sim}\left({Z}^{(r)}_{i}, {Z}^{(s)}_{i}\right) / \tau\right)}{\sum_{k=1}^{2L} \mathbf{1}_{k \neq i} \exp\left(\operatorname{sim}\left({Z}^{(r)}_{i}, {Z}_{k}\right) / \tau\right)},
\end{equation}
\noindent where $\tau$ is the temperature parameter, $Z^{(r)}_{i}$ and $Z^{(s)}_{i}$ denote the two augmented representations of the $i$-th graph in the batch (i.e., the `r' = `reconstruction' and `s' = `subgraph' versions) respectively, and $Z_k$ denotes the representation of the $k$-th graph in a set consisting of $ 2L $ representations, with $ L $ of $Z^{(r)}_i $'s and $ L $ of $ Z^{(s)}_i$'s. The function $\operatorname{sim}(\cdot, \cdot)$ calculates the cosine similarity between two representations, 
and the indicator function ${1}_{k \neq i}$ ensures that the similarity of a representation with itself is excluded from the denominator.

Finally, the overall loss function of our \model is:
{\color{blue}
\begin{equation}
\color{blue}
L_{\text{ALL}} = \lambda_{\text{rec}} L_{\text{rec}} + \lambda_{\text{JSE}} L_{\text{JSE}} + \lambda_{\text{CL}} L_{\text{CL}}.
\end{equation}
}
{\color{blue}
where $\lambda_{\text{rec}}, \lambda_{\text{JSE}}, \lambda_{\text{CL}}$ are scalar weights for the reconstruction loss $L_{\text{rec}}$, subgraph loss $L_{\text{JSE}}$, and contrastive loss $L_{\text{CL}}$, respectively. 
Each loss plays a complementary role: $L_{\text{rec}}$ preserves structural integrity under masking and denoising, $L_{\text{JSE}}$ guides the interpreter to extract concise, task-aligned subgraphs, and $L_{\text{CL}}$ sharpens class-discriminative geometry of the learned representations. 
To avoid introducing additional tuning, we fix $\lambda_{\text{rec}}{=}\lambda_{\text{JSE}}{=}\lambda_{\text{CL}}{=}1$ throughout, and Appendix~A demonstrates that this equal-weight setting is effective in our experiments.
}

\section{EXPERIMENTS}

\subsection{Experimental Setup}
{\color{blue}
To evaluate model performance, we conduct experiments on four publicly available ECG datasets—MITBIH, PTB-XL, ST-T, and the contemporary 12-lead Chapman–Shaoxing dataset—covering diverse lead configurations, sampling rates, durations, and diagnostic targets. Using the criteria given by the American Association for the Advancement of Medical Instrumentation (AAMI), we divide ECG signals into five groups: non-ectopic (N), supraventricular (S), ventricular (V), fused (F), and unknown (Q). For the Chapman–Shaoxing dataset, the original label space comprises numerous expert diagnostic statements. The distribution is markedly long-tailed, with many labels appearing in only a single or a few records. To obtain reliable estimates and emphasize clinically risk-relevant abnormalities, we aggregate diagnostic statements into five multi-label disease superclasses—\emph{Atrial Abnormalities}, \emph{Junctional Arrhythmias}, \emph{Ventricular Arrhythmias}, \emph{Conduction Abnormalities}, and \emph{Myocardial Ischemia/Infarction}—so that each recording may belong to multiple superclasses. Accordingly, on Chapman–Shaoxing we report results under the \emph{Anomaly Risk Categories} setting.
The dataset statistics are summarized in Table~\ref{tab:stat}.
}


\begin{table}[t]
\centering
\caption{Statistics of the ECG datasets used in this study.}
\label{tab:stat}
\begin{tabular}{lcccc}
\hline
Dataset & Samples & Leads & Signal length & Frequency\\
\hline
MITBIH   & 100687 & 2  & 9--60\,s & 360\,Hz \\
PTB-XL   & 21799  & 12 & 10\,s    & 500\,Hz \\
ST-T     & 791489 & 2  & 2\,h     & 250\,Hz \\
\textcolor{blue}{Chapman--Shaoxing} & \textcolor{blue}{45152} & \textcolor{blue}{12} & \textcolor{blue}{10\,s} & \textcolor{blue}{500\,Hz}\\

\hline
\end{tabular}
\end{table}

\begin{table*}[t]\small
    \centering
    \caption{Performance comparison of classification in the overall categories on the three ECG datasets. The best and second-best results in each column are marked in {\bf bold} and \underline{underlined}, respectively.  `--' indicates that we cannot find or reproduce the results due to private implementation of the original papers or inapplicable settings.}
    \begin{tabular}{cccccccccccccccccc}
        \hline
        \multirow{2}{*}{Method} & \multicolumn{4}{c}{MITBIH} & \multicolumn{4}{c}{ST-T} & \multicolumn{4}{c}{PTB-XL} \\
        & ACC & F1 & Sen  & Spe& ACC  & F1 & Sen  & Spe& ACC  & F1 & Sen  & Spe  \\
        \hline
                GAT~ & 0.9507 & 0.5667 & 0.5003& 0.6538 & 0.9910 & 0.1991 & 0.2000 & 0.4989  & 0.6168 & 0.4472 &0.4466 &0.5756 \\
        GIN~ & 0.9003 & 0.2663 & 0.2490  & 0.5355  
        & 0.9910  & 0.1991& 0.2000 & 0.4990
        & 0.4624 & 0.1265 & 0.2000  &0.4355 \\
        SAGPool & 0.9777 &0.7236 &0.6888 & 0.7050 &0.9914 &0.1991&0.2000 &0.4990 & 0.6809 & 0.5643 & 0.5562 &0.6171\\
        \hline
        MINIROCKET~ & \underline{0.9913} & 0.6601 & 0.7489 & 0.9274 & 0.9967 & 0.7264 & 0.5741 & 0.9274  & 0.7217 & 0.6179 &0.6048 & 0.9201  \\
        ResNet50~ & 0.9909 & 0.7629 & 0.7589&0.9886 & 0.9971 & 0.6757&0.6528&0.9568 & 0.6794  & 0.5631 & 0.5510 & 0.9085\\
        LSTM-FCN~ & 0.9684 & 0.7604 & 0.7524 & 0.7187 & 0.9790 & 0.2130 & 0.2139 & 0.8182 & \underline{0.7289} & \underline{0.6250} & 0.6016 & \underline{0.9221}\\

        TCN~ & 0.9008 & 0.2835 & 0.2549  & 0.5685 
            & 0.9913 &0.1991&0.2000 &0.4990 
            & 0.4635& 0.2369 &0.2524  & 0.4851 \\
        RCNNs & 0.9846 & 0.7196 & 0.7233& 0.9873 & -- & --& --& -- & -- & --& --& --\\
        CascadeCNN & 0.9907 &\underline{0.7665} & 0.7548&0.9874 &\underline{0.9972} & \underline{0.7479}& \underline{0.7029} & 0.9520& 0.6697 & 0.5752 &0.5803 & 0.9096\\
        Transformer &0.9912 &0.7633 &\underline{0.7600} & \underline{0.9895} & 0.9965&0.6333 & 0.6147& \underline{0.9575}& 0.7106&0.6028 &0.5959 &0.9206 \\
        \hline
{\color{blue}BMIRC} & {\color{blue}0.9849} & {\color{blue}0.6743} & {\color{blue}0.7533} & {\color{blue}0.9891} & {\color{blue}0.9922} & {\color{blue}0.6450} & {\color{blue}0.6401} & {\color{blue}0.9015} & {\color{blue}0.7255} & {\color{blue}0.6231} & {\color{blue}\underline{0.6102}} & {\color{blue}0.8942} \\
{\color{blue}WResHDual} & {\color{blue}0.9901} & {\color{blue}0.7622} & {\color{blue}0.7505} & {\color{blue}0.9875} & {\color{blue}0.9968} & {\color{blue}0.7403} & {\color{blue}0.6955} & {\color{blue}0.9490} & {\color{blue}0.7210} & {\color{blue}0.6190} & {\color{blue}0.5983} & {\color{blue}0.9195} \\

         \hline
          TCL & 0.9812 &0.7280 & 0.7084&0.9736&0.9951 & 0.3674&0.3536 &0.8990& 0.4546 & 0.1250 &0.2000 & 0.4214\\
          CPC & 0.5124 &0.1651 &0.1982&0.7989 & 0.8622&0.1874&0.1992  & 0.7993& 0.4546 & 0.1250 &0.2000 & 0.4214 \\
        \model (ours) & \textbf{0.9916} & \textbf{0.7712} & \textbf{0.7656} & \textbf{ 0.9910}& \textbf{0.9975}& \textbf{0.7719}& \textbf{0.7042}& \textbf{0.9592}& \textbf{0.7313} & \textbf{0.6400} & \textbf{0.6284} & \textbf{0.9248}\\
        \hline
    \end{tabular}

    \label{tab:Baseline}
\end{table*}
        
\begin{table*}[t]\small
    \centering
    \caption{Performance comparison of classification in the anomaly risk categories. The best and second-best results in each column are marked in {\bf bold} and \underline{underlined}, respectively.  `--' indicates that we cannot find or reproduce the results due to private implementation of the original papers or inapplicable settings.}
    \begin{tabular}{cccccccccccccccccc}
        \hline
        \multirow{2}{*}{Method} & \multicolumn{3}{c}{MITBIH} & \multicolumn{3}{c}{ST-T} & \multicolumn{3}{c}{PTB-XL} \\
        & ACC & F1 & Sen  & ACC  & F1 & Sen & ACC  & F1 & Sen \\
        \hline
        MINIROCKET~ &0.9230  & \underline{0.5702}& 0.6863&0.6339 &0.4605 &0.4676 & 0.5940&0.4350&0.5325\\
        LSTM-FCN~ &0.8441 &0.4867 &0.5792& 0.0474&0.0100 &0.0397 &\underline{0.6140}&\underline{0.4475}&\underline{0.5564 } \\

        CascadeCNN &0.9397 &0.5653 &0.6931 & \underline{0.8340}& 0.4705& \underline{0.6145} &0.5404 &0.4142 & 0.5193& \\
        Transformer&\underline{0.9434}&0.5642&\underline{0.7008}&0.7552&0.4336&0.5187&0.5884&0.4289&0.5248 \\
         \hline

{\color{blue}BMIRC} & {\color{blue}0.9385} & {\color{blue}0.5670} & {\color{blue}0.6955} & {\color{blue}0.8310} & {\color{blue}\underline{0.4725}} & {\color{blue}0.6120} & {\color{blue}0.6110} & {\color{blue}0.4450} & {\color{blue}0.5530} \\
{\color{blue}WResHDual} & {\color{blue}0.9355} & {\color{blue}0.5633} & {\color{blue}0.6910} & {\color{blue}0.8290} & {\color{blue}0.4690} & {\color{blue}0.6080} & {\color{blue}0.6085} & {\color{blue}0.4421} & {\color{blue}0.5501} \\
\hline
          TCL & 0.8682 &0.5300 & 0.6369&0.4851 & 0.1679&0.1920 & --&  -- &  -- \\
          CPC & 0.1025& 0.0278& 0.0233&0.0017 &0.0021 &0.0015 & --&  -- &  --\\
        \model (ours) & \textbf{0.9554}&\textbf{0.5721}&\textbf{0.7080}&\textbf{0.8472}&\textbf{0.6241}&\textbf{0.6896}& \textbf{0.6179}&\textbf{0.4712} &\textbf{0.5688} \\
        \hline
    \end{tabular}

    \label{tab:Baseline_abnomal}
\end{table*}

\noindent \textbf{MITBIH}: The MITBIH dataset~\cite{moody2001impact} contains 48 half-hour two-channel ECG recordings sampled at 360 Hz.
Four ECG records of pacemaker users (numbers 102, 104, 107, and 217) were deleted following the standards proposed by AAMI.
As in the previous work~\cite{sellami2019robust,atal2020arrhythmia,hammad2019novel}, we use the single lead and modified limb lead ll (MLII) as the input of the baselines.

\noindent \textbf{PTB-XL}: The PTB-XL dataset~\cite{wagner2020ptb} contains 21,837 12-lead ECG records sampled at 500 Hz for exactly 10 seconds in duration. We pre-process this dataset following the AAMI criteria, resulting in 21,799 ECG recordings with 12 leads, partitioned by PTB-XL superclass labels.


\noindent \textbf{ST-T}: The European ST-T dataset~\cite{taddei1992european} contains 90 excerpts of annotated Holter recordings from 79 subjects. Each recording has a duration of two hours and consists of two signals, and each is sampled at 250 samples per second. We divide the two hours of recordings by heartbeat labels.

\noindent \textcolor{blue}{\textbf{Chapman–Shaoxing}: The Chapman-Shaoxing dataset~\cite{zheng2022large} contains 45,152 10-second, 12-lead ECGs at 500 Hz, labeled by licensed physicians with diagnostic statements. Labels are mapped to Systematized Nomenclature of Medicine—Clinical Terms (SNOMED CT) for multi-label classification, and records are provided in WaveForm Database (WFDB) with per-record metadata, offering a large and contemporary dataset for arrhythmia modeling.}

\subsection{Implementation Details} 
We randomly allocate 70$\%$ of the ECG signals to the training set and retain the remaining 30$\%$ for the test set, ensuring robust evaluation of the model's performance on unseen data. \eat{To adapt the baseline model to ECG signal lengths, we employ a sliding window to partition the original ECG signal into fixed-length segments, aligning with the model's input expectations.} 
Hyperparameters are heuristically set with a threshold parameter $\Theta$ filtering out 75\% of graph connections, \(\delta = 0.80\) to control edge retention, a masking rate $\rho$ = 70\%, and a scaling factor $\gamma$ of $2$ to balance reconstruction errors and enhance generalization. The default batch size is set to 1024, and model training is conducted on an NVIDIA GeForce RTX 4090. In the cases where GPU memory limits are encountered, we appropriately reduce the batch size for training purposes. In the downstream tasks, we use an MLP to classify the ECG representation. For the interpretation tasks, we directly utilize the trained Feature Subgraph Module to perform subgraph extraction and provide interpretability. We use several unified metrics to assess the 
model performance, including accuracy, F1 score, specificity, and sensitivity. These metrics are computed using the macro method to identify the advantages and areas for improvement in cardiac arrhythmia detection and heartbeat classification methods.

\subsection{Comparison with Baseline Methods}

We evaluate the effectiveness of our proposed \model and compare with 12 ECG classification baselines, including GNN-based models (GAT \cite{velickovic2017graph}, GIN~\cite{xu2018powerful}, and SAGPool~\cite{lee2019self}),  CNN-based models (MINIROCKET~\cite{dempster2021minirocket}, LSTM-FCN~\cite{karim2017lstm}, ResNet50~\cite{he2016deep}, TCN~\cite{bai2018empirical},  RCNNs~\cite{le2021multi}, CascadeCNN~\cite{li2015convolutional}, and Transformer~\cite{vaswani2017attention}), and self-supervised learning models (TCL~\cite{poppelbaum2022contrastive} and CPC~\cite{oord2018representation}). {\color{blue}To provide a contemporary comparison in ECG-specific modeling, we additionally include two recent, ECG-tailored baselines: a multimodal fusion method that combines spatiotemporal and frequency representations (WResHDual)~\cite{liu2025multimodal}, and a bimodal masked autoencoder with internal representation connections (BMIRC)~\cite{wei2025bimodal}.}
Since the baselines have various input dimension requirements, we resample the datasets in order to meet the input requirements of each model. We experiment with classification performance in both the overall categories and risk categories. The results show that our \model model not only adapts to the diverse nature of ECG data but also excels in identifying the nuanced variations of risk categories, demonstrating the robustness and adaptability of our new method.

\textbf{Classification Performance in the Overall Categories:}
As shown in Table~\ref{tab:Baseline}, \textbf{in the realm of GNN classification models}, GNNs exhibit commendable performance on the relatively simple MITBIH dataset. However, their effectiveness diminishes on the significantly imbalanced ST-T dataset, on which the message-passing technique struggles to differentiate among fewer ECG signal categories. \textbf{CNN-based models} demonstrate strong performance on the datasets aligned with their input structure but struggle with variability in ECG data, as they are unable to maintain high performance across different datasets. RCNNs classify arrhythmias by transforming ECG signals into image data, and consequently face scalability issues when applied to large-scale ECG datasets. Due to the significant sizes and complexity of the ST-T and PTB-XL datasets, training the model on them is not feasible with such an approach. Therefore, results in certain metrics cannot be computed, and we use `--' as a placeholder to indicate the missing values in Table~\ref{tab:Baseline}. \textbf{In self-supervised learning models,} it is observable from Table~\ref{tab:Baseline} that TCL achieves better performance on short ECG sequences but fails to represent effectively on the longer sequence PTB-XL dataset. CPC shows poor classification performance on the three ECG datasets. 
This can be attributed to the periodicity inherent in ECG signals, which poses a challenge to the intrinsic predictive feature learning method employed by CPC. Consequently, this approach struggles to effectively capture and represent key discriminative features necessary for accurate classification of ECG signals.
From Table~\ref{tab:Baseline}, it is evident that no single known method dominates as the runner-up (to our \model) across the different datasets. In comparison, our \model model captures a unified representation of ECG signals across these datasets, avoiding classification performance degradation typically caused by dataset format variations. This demonstrates the robustness and adaptability of our model in achieving high classification accuracy, regardless of the dataset complexities.

\textbf{Classification Performance in the Risk Categories:} Due to the predominance of the normal categories in general ECG data, the results obtained for the overall classification performance can obscure the models' ability to classify abnormal categories. To further accurately evaluate the models' performance in the anomaly risk categories, we compare our \model with those baselines yielding second-best performance and the self-supervised learning baselines.  Table~\ref{tab:Baseline_abnomal} shows the results. It is important to note that we omit the results of specificity metrics because our focus is on the performance of the abnormal categories. On the PTB-XL dataset, TCL and CPC cannot effectively distinguish between normal and abnormal samples, and hence we use `--' to indicate this. The results reveal significant performance disparities between normal and abnormal categories across the baseline models, as demonstrated in Tables \ref{tab:Baseline} and \ref{tab:Baseline_abnomal}.  \model consistently excels in classifying abnormal samples, demonstrating its ability to accurately identify abnormal patterns in ECG data. This superior performance underscores the effectiveness and reliability of \model in detecting critical cardiac abnormalities.


\textcolor{blue}{
\textbf{Classification Performance on the Multi-Label Dataset:}
As summarized in Table~\ref{tab:shaoxing_results}, \model\ attains the strongest accuracy, specificity, and AUC on Chapman--Shaoxing, while remaining competitive on F1 and precision. Relative to WResHDual, \model\ accepts a slight trade-off in F1 and precision for clearer gains in separability and in true-negative control, yielding an operating profile that is robust to threshold selection—particularly desirable for multi-label clinical screening. CNN baselines such as ResNet and Transformer show strong specificity but lower separability and overall accuracy and F1 compared with \model. GNN baselines (GAT, GIN, SAGPool) group at a lower performance tier across metrics, and methods like MINIROCKET and BMIRC underperform broadly. Overall, Table~\ref{tab:shaoxing_results} indicates that \model\ combines high separability and specificity with competitive positive-call performance, offering a balanced choice for contemporary multi-label ECG analysis.}

\begin{table}[h!]
\centering
\caption{{\color{blue}Detailed Performance on the Chapman--Shaoxing Dataset.}}
\label{tab:shaoxing_results}
\small
\begin{tabular}{lccccc}
\hline
\multirow{2}{*}{\textbf{\color{blue}Method}} & \multicolumn{5}{c}{\textbf{\color{blue}Chapman--Shaoxing}} \\
& \textbf{\color{blue}ACC} & \textbf{\color{blue}F1} & \textbf{\color{blue}Precision} & \textbf{\color{blue}Spe} & \textbf{\color{blue}AUC} \\
\hline
\color{blue}GAT           & {\color{blue}0.574} & {\color{blue}0.570} & {\color{blue}0.520} & {\color{blue}0.908} & {\color{blue}0.852} \\
\color{blue}GIN           & {\color{blue}0.597} & {\color{blue}0.573} & {\color{blue}0.540} & {\color{blue}0.921} & {\color{blue}0.857} \\
\color{blue}SAGPool       & {\color{blue}0.583} & {\color{blue}0.570} & {\color{blue}0.530} & {\color{blue}0.911} & {\color{blue}0.853} \\
\hline
\color{blue}MINIROCKET    & {\color{blue}0.465} & {\color{blue}0.460} & {\color{blue}0.400} & {\color{blue}0.844} & {\color{blue}0.751} \\
\color{blue}ResNet50      & {\color{blue}0.640} & {\color{blue}0.682} & {\color{blue}0.630} & {\color{blue}0.939} & {\color{blue}\underline{0.930}} \\
\color{blue}LSTM-FCN      & {\color{blue}0.597} & {\color{blue}0.584} & {\color{blue}0.530} & {\color{blue}0.916} & {\color{blue}0.857} \\
\color{blue}CascadeCNN    & {\color{blue}0.632} & {\color{blue}0.657} & {\color{blue}0.621} & {\color{blue}0.923} & {\color{blue}0.901} \\
\color{blue}Transformer   & {\color{blue}0.641} & {\color{blue}0.622} & {\color{blue}0.590} & {\color{blue}\underline{0.939}} & {\color{blue}0.891} \\
\hline
\color{blue}{BMIRC}       & {\color{blue}0.453} & {\color{blue}0.471} & {\color{blue}0.430} & {\color{blue}0.853} & {\color{blue}0.773} \\
\color{blue}{WResHDual}   & {\color{blue}\underline{0.661}} & \textbf{\color{blue}{0.730}} & \textbf{\color{blue}{0.710}} & {\color{blue}{0.923}} & {\color{blue}{0.926}} \\
\hline
\textbf{\color{blue}VARS (ours)} & \textbf{\color{blue}{0.669}} & {\color{blue}\underline{0.728}} & {\color{blue}\underline{0.690}} & \textbf{\color{blue}{0.945}} & \textbf{\color{blue}{0.939}} \\
\hline
\end{tabular}
\end{table}

\begin{figure*}[!htbp]
    \centering
    \includegraphics[width=0.98\textwidth]{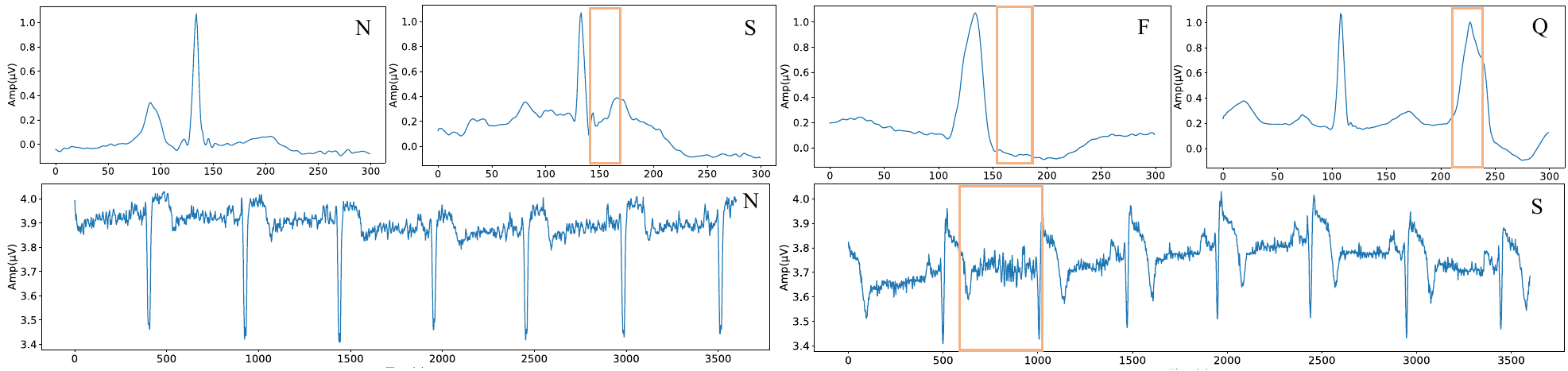}
    \caption{\textbf{Visualization of ECG interpretability.} The top portion is an interpretable demonstration at the heartbeat structure level, and the bottom portion is an interpretable demonstration at the heartbeat level.}
    \label{fig:Figure 2}
\end{figure*}

\subsection{Model Interpretability}

The interpretability of the \model model enables it to provide meaningful explanations for its classification decisions.
Traditional interpretability often offers only a characteristic explanation of the ECG signal, which can be perplexing. To address this issue, we implement an interpretable method that integrates both qualitative and quantitative approaches, aligning with experts' practice perspectives.
\subsubsection*{Qualitative Analysis}
We conduct granular model interpretability studies using data from the MITBIH and PTB-XL datasets. The results are presented in Figure \ref{fig:Figure 2}. We investigate two different granularity levels of interpretability: the heartbeat structure level and heartbeat level. \model allows medical professionals to make ECG-interpretable fine-grained choices based on their diagnostic needs to ensure that they can quickly locate abnormal ECG segments for ECG-aided diagnosis. In Figure \ref{fig:Figure 2}, we see various labels such as N, S, F, and Q, which correspond to normal heartbeats, supraventricular premature beats, atrial fibrillation, and unknown rhythms, respectively. Their specific manifestations in the ECG data are clearly illustrated. In the heartbeat structure level, as shown in the top portion of Figure \ref{fig:Figure 2}, the leftmost part illustrates a schematic of a normal heartbeat category, while the other three parts display schematics of ECG heartbeats for three different disease types. \model can swiftly delineate the structure of an abnormal ECG segment to assist doctors in rapid diagnosis. We further consulted medical experts, who confirmed that the delineated abnormal regions are consistent with clinical diagnostic standards. Within the scope of the heartbeat level, the bottom portion of Figure~\ref{fig:Figure 2} offers an interpretable analysis within the continuous ECG signal. The left part gives a normal ECG signal, and the right part illustrates the range of delineated abnormality classes. Doctors can select one or more heartbeats to allow the model to determine the range, thereby facilitating rapid diagnosis.

\begin{figure}[t]
  \centering
  \subfloat[]{\includegraphics[width=0.3\linewidth]{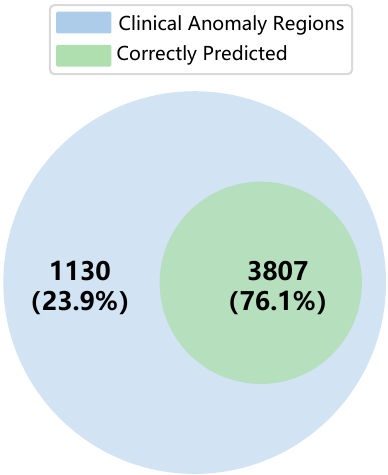}\label{fig:match_combo_a}}
  \hspace{0.15\linewidth}%
  \subfloat[]{\includegraphics[width=0.47\linewidth]{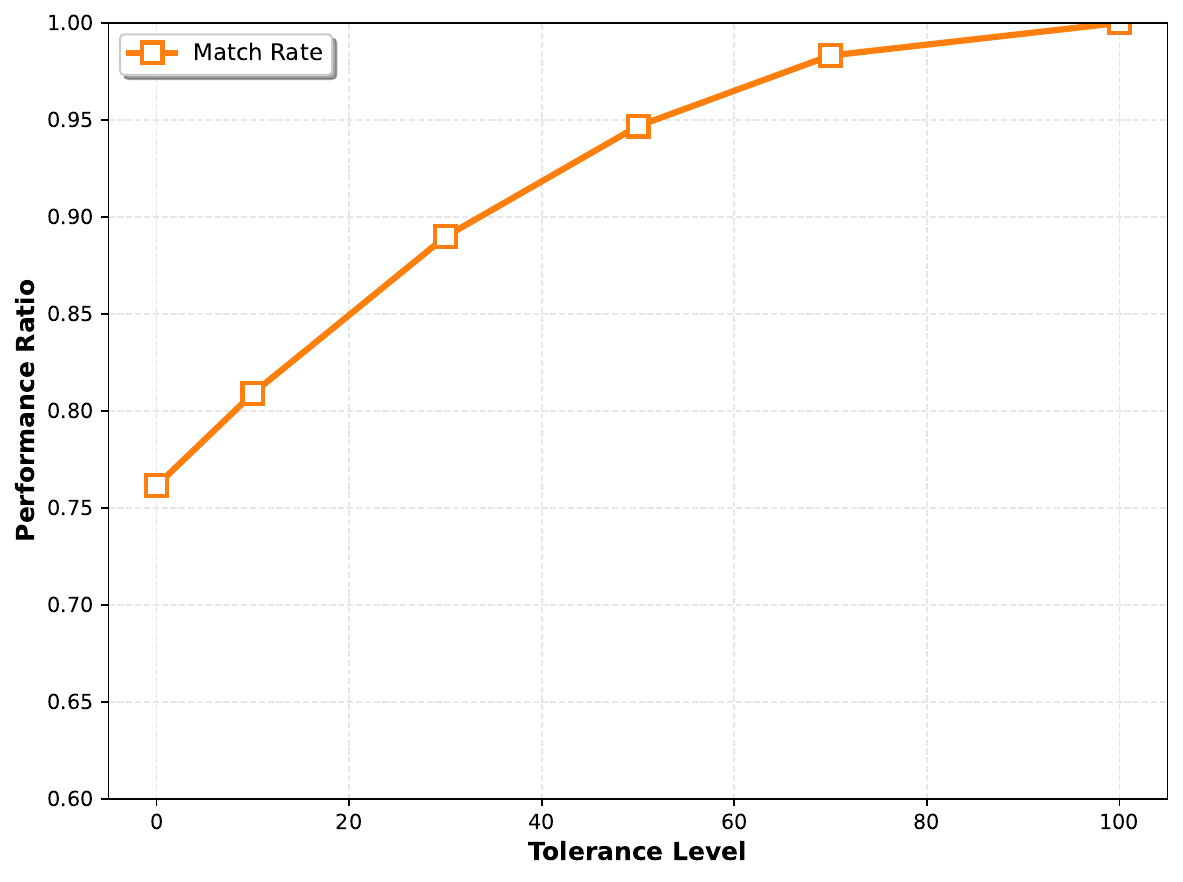}\label{fig:match_combo_b}}
  \caption{{\color{blue}\textbf{Match between clinical annotations and interpreter outputs.}
  (\textbf{a}) Venn diagram on the 5{,}000-record set showing where the interpreter’s top-1 segment overlaps the clinically annotated interval.
  (\textbf{b}) Match-rate curve as the window tolerance increases, showing a smooth rise that indicates stable agreement between the interpreter’s top-1 segment and the clinical markings.}}
  \label{fig:match_combo}
\end{figure}

{\color{blue}
\subsubsection*{Quantitative Analysis}

\color{blue}We quantify interpretability using model-internal signals produced by the Feature Subgraph Module. For each sample, the module yields node- and edge-level importance on the ECG graph and highlights the time-interval segments that constitute the explanatory subgraph. We report compact summaries of these outputs via per-sample dashboards to complement the qualitative analyses in Figure~\ref{fig:Figure 2}. Full visualization layouts and the machine-readable summary schema are provided in Appendix B. 

To assess whether the interpreter’s evidence aligns with clinical reading, we curated a set of 5{,}000 ECG records and had clinical annotate a single diagnostically salient time-interval per record. Our interpreter then returns a \emph{top-1} segment (the highest-importance patch) per record. We count a prediction as a match when the interpreter’s top-1 segment overlaps the reference interval derived under standard clinical interpretation principles. Figure~\ref{fig:match_combo}(a) summarizes the overlap between the clinical reference annotations and the interpreter’s top-1 segments, while Figure~\ref{fig:match_combo}(b) shows how the match rate varies as the tolerance window increases. Together, these plots indicate that the interpreter’s most salient window aligns well with clinical markings and that the match rate increases smoothly as the tolerance broadens. We also provide a full 12-lead visualization where clinician-marked segments and the interpreter’s top-1 segments. The panel shows both normal and abnormal examples and reports the segment indices and time ranges for clarity. The full figure is included in Appendix~B.

}




\textcolor{blue}{
\subsection{Hyperparameter Sensitivity and Computational Efficiency}}

\begin{figure}[t]
  \centering
  \includegraphics[width=\linewidth]{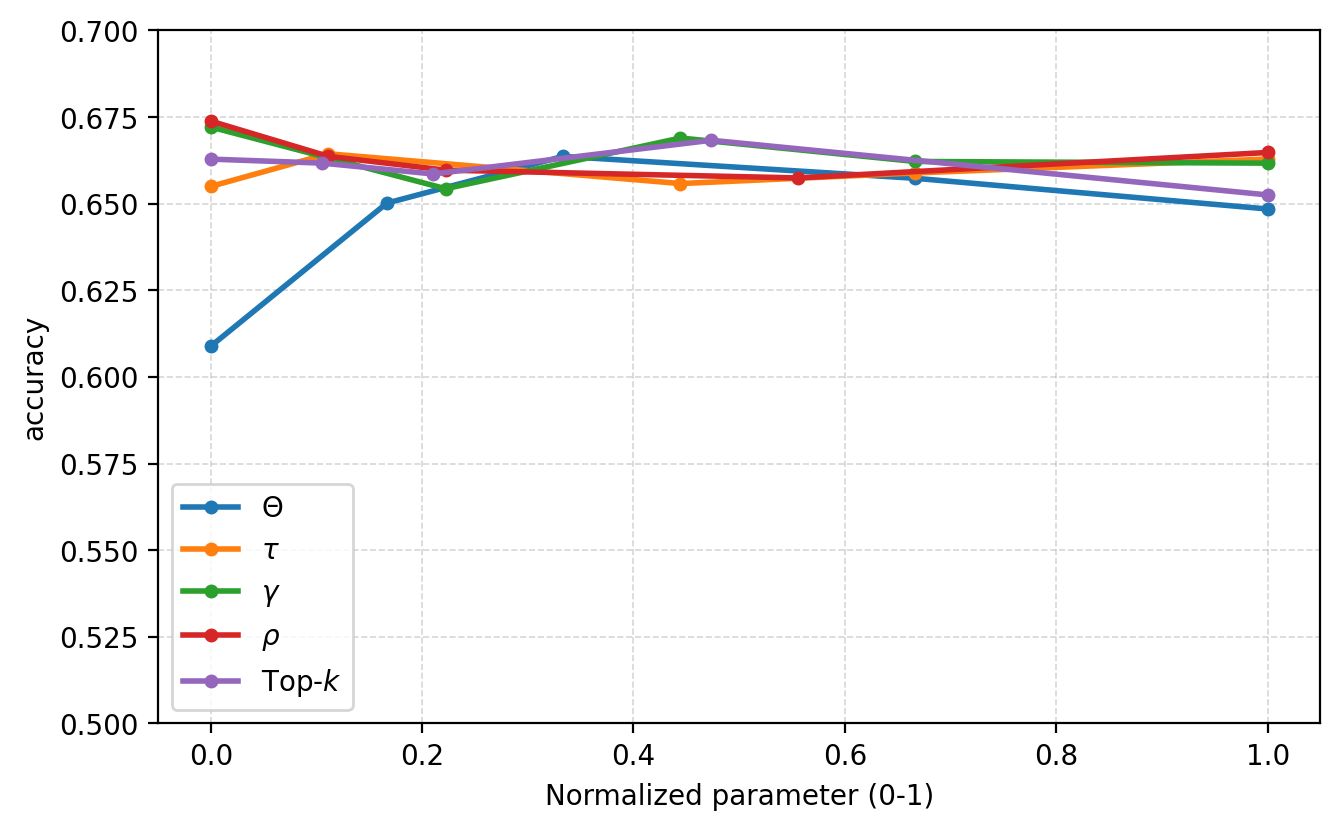}
  \caption{\color{blue}Parameter sensitivity of \model\ over five hyperparameters:
  $\Theta$, Top-$k$, masking rate $\rho$, $\gamma$, and $\tau$.
  \textcolor{blue}{Each curve uses a five-point grid centered at our default (the \emph{third} tick), with two symmetric values on each side.
 }}
  \label{fig:sensitivity}
\end{figure}

\textcolor{blue}{To assess robustness, we conduct a one–factor–at–a–time sensitivity study on five hyperparameters—graph threshold $\Theta$, Top-$k$, masking rate $\rho$, reconstruction scaling $\gamma$, and contrastive temperature $\tau$. For each hyperparameter, we evaluate five values by centering the grid at our model’s setting (the third point, median of the grid) and expanding symmetrically to both sides. The results show stable performance over broad ranges in Figure~\ref{fig:sensitivity}. The sensitivity curves remain stable around the default setting, and the observed variations across the five ablations are modest, indicating that the default configuration lies on a broad performance plateau and serves as a reliable choice without heavy per-dataset retuning.} 

\textcolor{blue}{In addition, we report the computational profile (parameter, FLOPs and latency). As summarized in Table~\ref{tab:complexity}, \model\ maintains a moderate parameter count among ECG-specific classifiers and achieves low FLOPs with competitive latency, and while smaller than large ECG models, it still delivers faithful, clinically meaningful explanations. Notably, although \model is slightly larger than lightweight CNNs, it keeps inference times short enough for real-time use and sustains stable performance across heterogeneous settings. } 

\begin{table}[h]
  \centering
\caption{{\color{blue}Computational complexity summary. Metrics reported: parameters (millions), FLOPs (billions), and end-to-end latency (ms per sample) on an RTX\,4090 (lower is better). Best in bold; second best underlined.}}
  \label{tab:complexity}
  \begin{tabular}{lccc}
    \hline
    \textbf{\textcolor{blue}{Model}} & \textbf{\textcolor{blue}{Model Size (M)}} & \textbf{\textcolor{blue}{FLOPs (G)}} & \textbf{\textcolor{blue}{Latency (ms)}} \\
    \hline
    \textcolor{blue}{BMIRC}         & \textbf{\textcolor{blue}{19.38}} & \underline{\textcolor{blue}{7.6}}  & \textbf{\textcolor{blue}{5.25}} \\
    \textcolor{blue}{WResHDual}     & \textcolor{blue}{33.08} & \textcolor{blue}{16.53} & \textcolor{blue}{13.11} \\
    \textbf{\textcolor{blue}{\model\ (ours)}} & \underline{\textcolor{blue}{23.39}} & \textbf{\textcolor{blue}{5.15}} & \underline{\textcolor{blue}{6.89}} \\
    \hline
  \end{tabular}
\end{table}
\subsection{Ablation Studies}

 We validate the efficacy of each module within our model through ablation experiments. Specifically, we remove the following components to examine their individual impact on performance:  the graph structure (G-S), denoising reconstruction module (D-R),  and graph-based contrastive learning (G-C), as well as the three related loss functions. The results are presented in Table~\ref{tab:adience ablation}. It is evident that the removal of the corresponding modules led to a decline in performance metrics to varying extents. This confirms the critical importance of these three modules in attaining model performance. The G-S and D-R modules, in particular, showed a significant enhancement in performance, suggesting that the construction of ECG graph representation and noise reduction reconstruction within the model are crucial components. The D-R module gives the most obvious improvement of the three parts, and the denoising reconstruction occupies a large part of the role, which can remove ECG noise signals and retain the essential characteristics.

Additionally, we show the effects of the three loss functions in Table~\ref{tab:adience ablation}. The removal of any loss function resulted in a certain degree of performance decrease, indicating that all these loss functions positively impact the model's performance. 
The complete model demonstrates superior performance across all the tests, illustrating the significance of our proposed modules and loss functions in enhancing ECG classification accuracy, enabling versatile and risk-sensitive cardiac diagnosis.

\begin{table}[t]
    \centering
    \caption{Results of ablation study
    on three different modules and three loss functions of our model.
    }\label{tab:adience ablation}
    \scalebox{0.9}{
    \begin{tabular}{c c c c c c}
        \hline
        Dataset      &      Module           & ACC              & F1 & Sen& Spe\\
        \hline
        \multirow{7}{*}{MITBIH}          &    w/o G-S                                      &0.9828& 0.7400 &0.7290& 0.9805\\
                    &    w/o D-R                   &  0.9775                & 0.7184 &0.7061&0.9755 \\
                    & w/o G-C                       & 0.9818         & 0.7269&0.7059&0.9803  \\

                 & w/o ${L_{r e c}}$              & 0.9757                & 0.7096 &0.6807  &0.9672\\
                &   w/o ${L_{JSE}}$                      & 0.9758        & 0.7102 & 0.6806 & 0.9663  \\
                &  w/o ${L_{CL}}$                           & 0.9761         & 0.7113& 0.6788  & 0.9659\\
                 & All            & \textbf{0.9916}         & \textbf{0.7712}   & \textbf{0.7656}   & \textbf{0.9910}  \\
        \hline

    \end{tabular}}
\end{table}

{\color{blue}
\section{Discussion}

Motivated by clinical use, we develop a unified and interpretable ECG representation that handles diverse acquisition configurations while providing clinician-ready evidence. \model converts heterogeneous waveforms into a graph representation so a single encoder can process different lead counts, sampling rates, and durations. The explanation pathway surfaces time-stamped intervals that support each prediction and can be overlaid in standard ECG viewers, enabling rapid triage and targeted review.
At inference, the runtime path is lean: attention-based graph construction, a single encoder pass, and on-demand evidence visualization that reuses encoder embeddings with modest overhead. As summarized in Table~\ref{tab:complexity}, \model offers a favorable compute profile—moderate parameter count, low FLOPs, and competitive end-to-end latency—while the interpretability results (Figures~3--4 and Appendix~B) show that highlighted intervals align well with clinically informed annotations. Together, these properties suggest that \model is a promising candidate for high-throughput ECG review in contemporary clinical workflows, and broader real-world validation remains an important next step.
}

\section{Conclusion}

In this work, we introduce the innovative \model model for a unified ECG signal representation, which effectively tackles the challenges posed by the heterogeneity of ECG signals and the deficiencies in risk signal detection due to sample imbalances. \model transforms ECG signals into versatile graph structures that capture essential diagnostic features, regardless of variations in lead count, sampling frequency, and duration. This graph-based approach enhances diagnostic sensitivity by enabling precise localization and identification of anomalous ECG patterns that typically elude conventional analysis methods. By integrating denoising reconstruction and contrastive learning, \model not only preserves the raw ECG information but also accentuates pathognomonic patterns. The experimental results demonstrate that \model consistently excels in performance across three widely-used datasets and shows significant improvements in detecting risk signals. Additionally, \model enhances the interpretability of classification results through various granular levels of analysis (discussed below), thereby assisting doctors in making informed decisions.

\textcolor{blue}{While this work establishes a strong foundation, the path to clinical deployment requires further investigation. Future work will focus on two key areas. First, a more extensive evaluation on real-world, noisy data from continuous monitoring devices is necessary to confirm the model’s robustness outside of curated datasets. Finally, prospective clinical trials are essential to validate the clinical efficacy of \model and to quantify its impact on diagnostic accuracy, workflow efficiency, and patient outcomes. We believe that addressing these challenges will pave the way for \model to become an invaluable tool for comprehensive cardiac health assessment.}

\bibliographystyle{IEEEtran}
\bibliography{sample-base}

\end{document}

\begin{figure*}[h]
  \centering
  \includegraphics[width=0.9\linewidth]{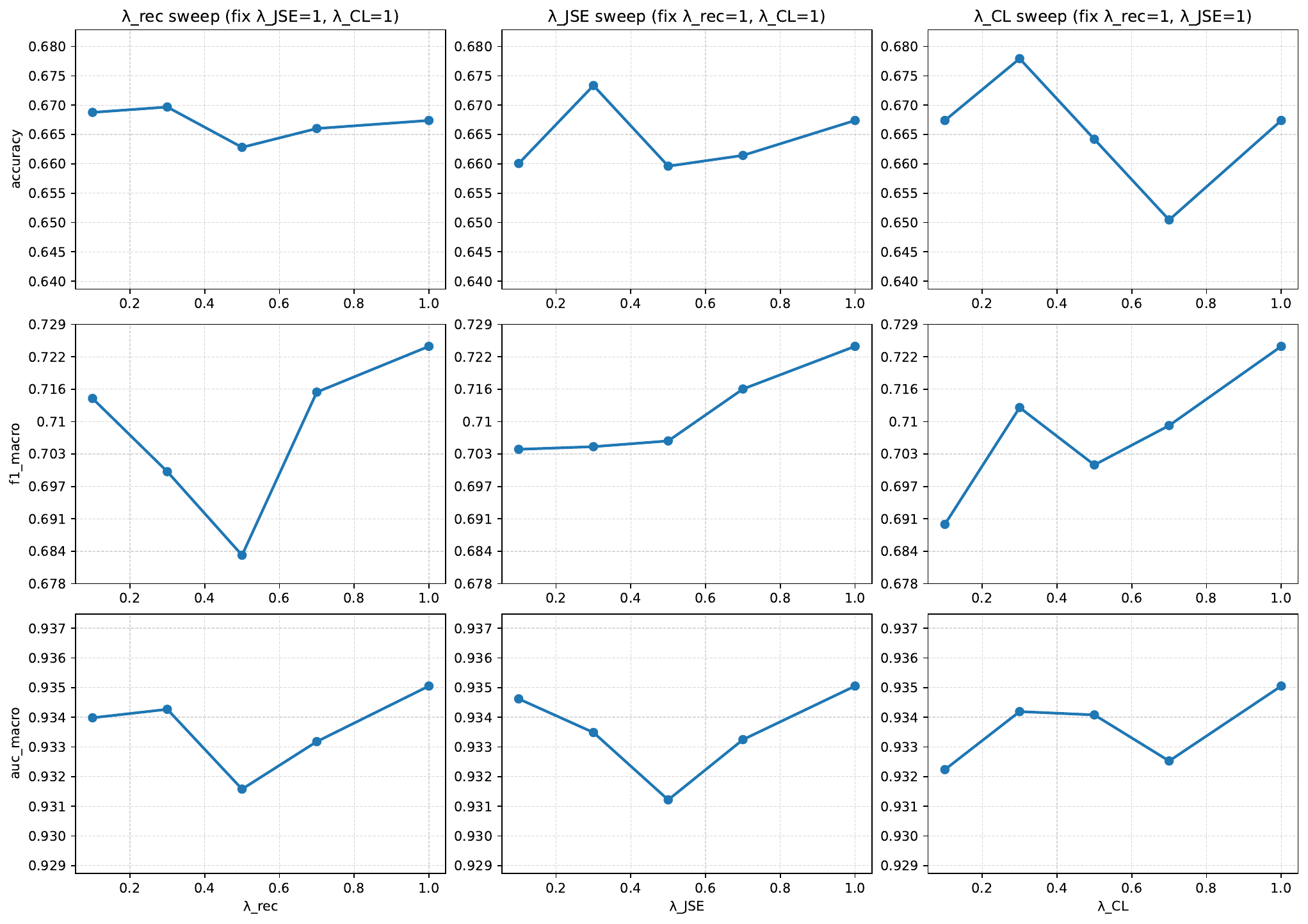}
  \caption{{\color{blue}One-factor-at-a-time sensitivity of loss weights. Columns sweep a single weight ($\lambda_{\text{rec}}$, $\lambda_{\text{JSE}}$, $\lambda_{\text{CL}}$) over $[0.1,\,1.0]$ with the other two fixed at $1$. Rows report accuracy, macro-F1, and macro-AUC. Curves are stable across the range, and the equal-weight point ($\lambda{=}1$) lies on a broad plateau across metrics.}}
  \label{fig:loss_sensitivity}
\end{figure*}

\appendices
{\color{blue}
\section*{Appendix A: Analysis of Loss Components}

We examine how the three objectives—reconstruction ($L_{\text{rec}}$), JSE ($L_{\text{JSE}}$), and contrastive loss ($L_{\text{CL}}$)—interact during training and how their relative weights influence downstream performance. To make the analysis transparent, we adopt a one-factor-at-a-time design: in each sweep we vary a single weight over $[0.1,\,1.0]$ while holding the other two weights at $1$. Concretely, we sweep $\lambda_{\text{rec}}$ with $\lambda_{\text{JSE}}{=}1$ and $\lambda_{\text{CL}}{=}1$, sweep $\lambda_{\text{JSE}}$ with $\lambda_{\text{rec}}{=}1$ and $\lambda_{\text{CL}}{=}1$, and sweep $\lambda_{\text{CL}}$ with $\lambda_{\text{rec}}{=}1$ and $\lambda_{\text{JSE}}{=}1$. All three sweeps are conducted on the Chapman--Shaoxing 12-lead dataset (multi-label) under the same preprocessing, train/test split, and training protocol described in the Experimental Setup, to reflect contemporary acquisition conditions. For each setting we report accuracy, macro-F1, and macro-AUC.

As shown in Figure~\ref{fig:loss_sensitivity}, two general patterns are apparent across the three sweeps: (i) the metric curves remain fairly flat over a broad range, with no consistent monotonic improvement when any single weight departs substantially from 1, and (ii) the vicinity of the equal-weight setting (\(\lambda_{\text{rec}}=\lambda_{\text{JSE}}=\lambda_{\text{CL}}=1\)) forms a wide performance plateau, where accuracy, macro-F1, and macro-AUC are near their maxima. Overall, these results indicate limited sensitivity to moderate reweighting of the three objectives and support the use of fixed equal weights for simplicity.

}

\begin{figure*}[t]
  \centering
  \includegraphics[width=0.98\textwidth]{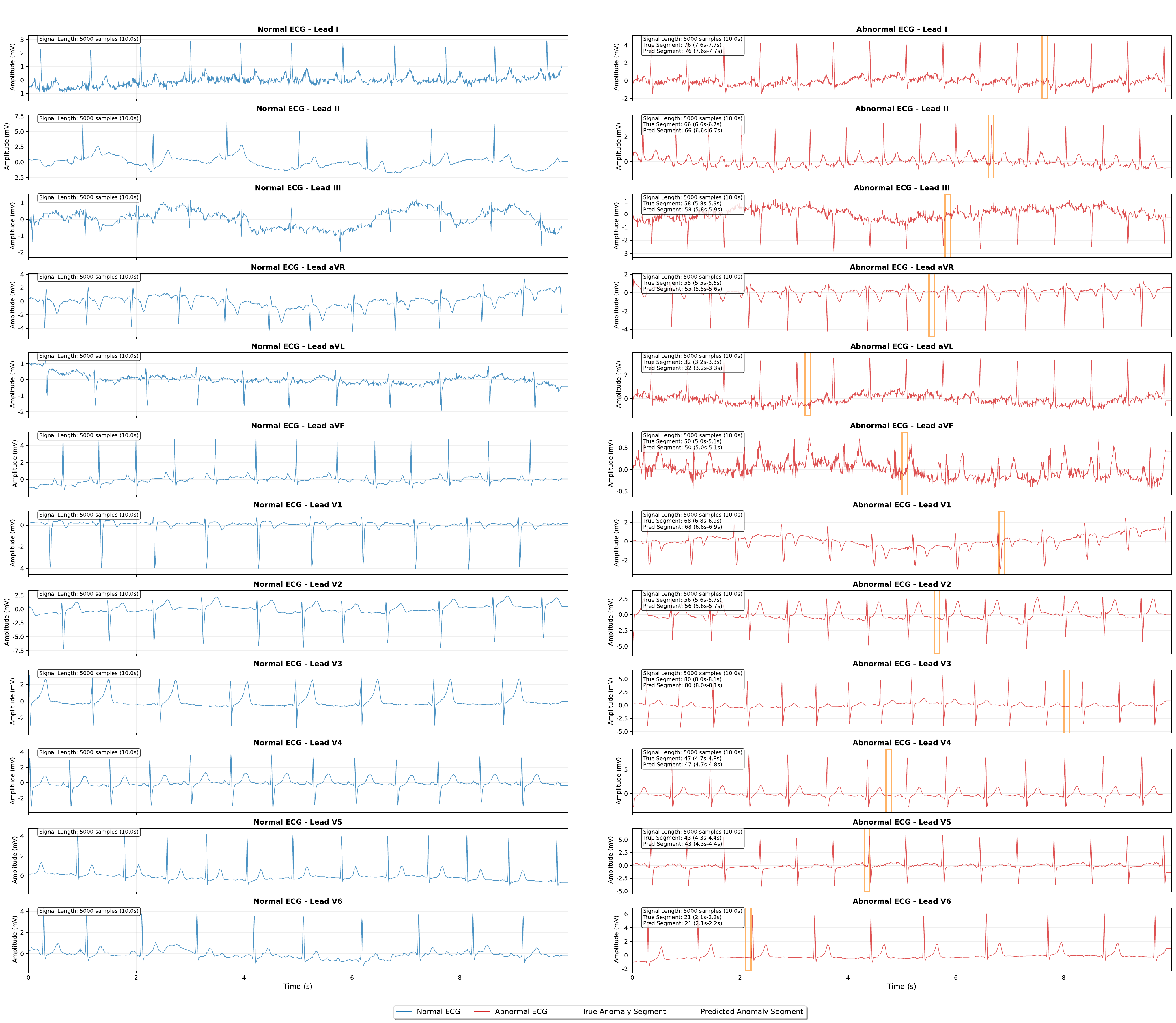}
\caption{\color{blue}\textbf{Lead-wise overlay of clinically annotated and model-interpreted segments}. \textbf{Left:} a normal ECG example (10\,s per lead). \textbf{Right:} an abnormal ECG example, where the interpreter highlights the most salient abnormal segment(s). The top-1 segment is indicated together with its index and time range.}
  \label{fig:full_signal_overlay}
\end{figure*}

{\color{blue}
\section*{Appendix B: Explainability Analysis and Visualization Details}

Consistent with the quantitative analysis in the main text, we quantify interpretability using model–internal signals emitted by the Feature Subgraph Module, without introducing an external post-hoc explainer. For each sample, the module produces node- and edge-level importance on the ECG graph and highlights the time-interval segments that constitute the explanatory subgraph. The composite Figure \ref{fig:appB1} below gathers these outputs into a single, per-sample dashboard to complement the qualitative plots in the main paper.

Operationally, ECG signals are split into fixed-size patches, and the module scores each patch to obtain patch importance. Saliency rendering uses a configurable threshold (default $\tau{=}0.3$), and top-$K$ or quantile selection can be used to control display sparsity. The per-sample dashboard can be produced for any correctly classified case (e.g., the highest-confidence instance within a class).

The dashboard comprises four coordinated panels generated directly from the Feature Subgraph Module: (i) A multi-lead ECG with salient intervals, where the waveform is uniformly partitioned into time patches and segments exceeding a user-set threshold are shaded. (ii) A node-importance heatmap, obtained by zero-padding and reshaping the node vector into a near-square matrix for inspection. (iii) An edge-importance histogram, providing a distributional view of relational weights with reference lines for quick calibration, indicating whether evidence concentrates on a few high-weight connections or is more diffuse, and (iv) a prediction summary showing class probabilities.

Figure~\ref{fig:full_signal_overlay} provides a lead-wise view for one normal case (left) and one abnormal case (right), each showing the full 10\,s waveform per lead as a complementary view. Each is labeled by its patch index and time range. In the normal example, no pathologic interval is marked, which is consistent with routine readings. In the abnormal example, the top-1 window selected by the interpreter aligns with the clinically annotated interval across multiple leads. This illustrates concordance between the model’s most salient evidence and clinical labeling. This full-signal overlay complements the per-sample dashboard by revealing how highlighted windows cohere across leads and by making disagreements—when present—visually apparent.

\begin{figure*}[t]
  \centering
  \includegraphics[width=\linewidth]{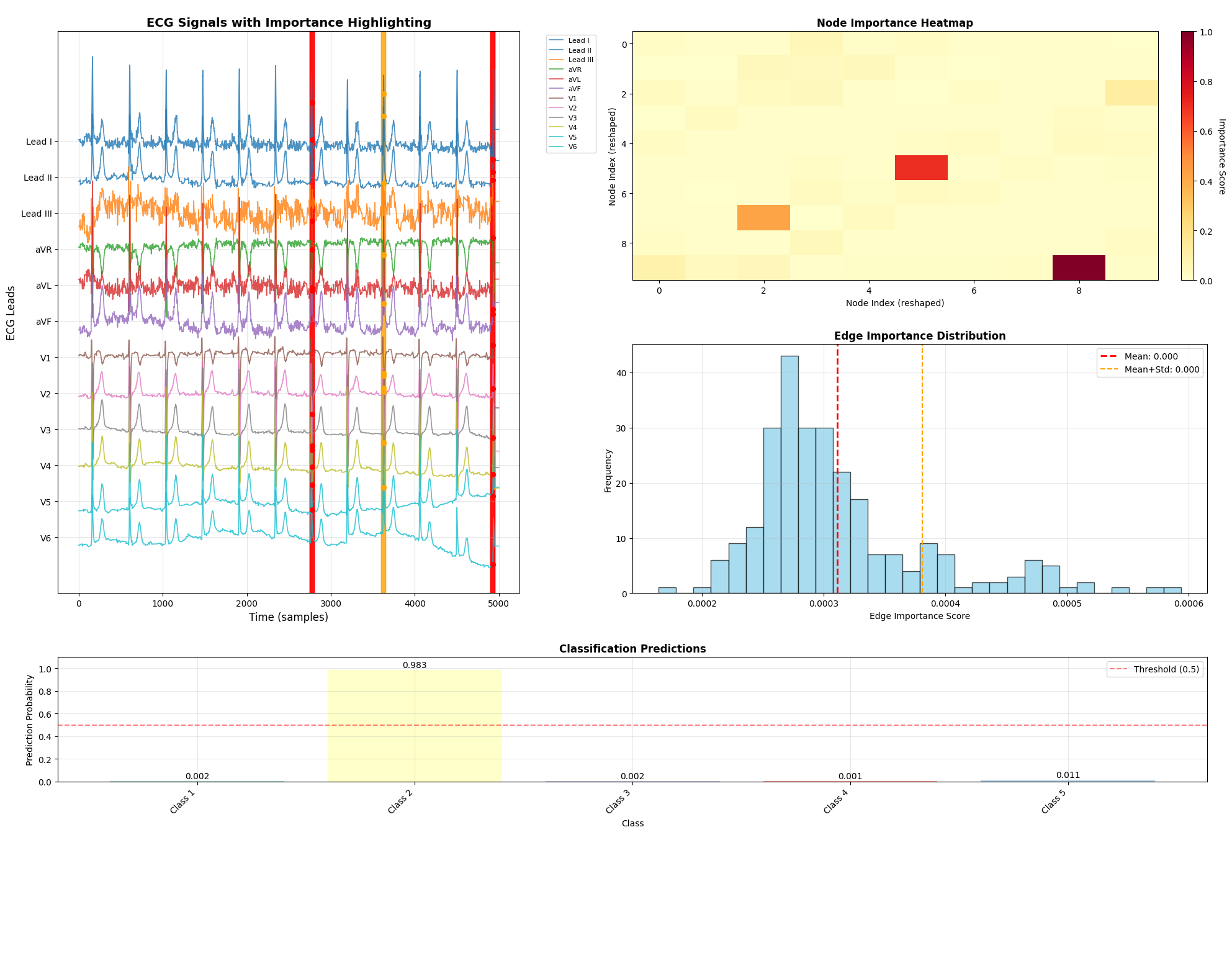}
  \caption{\color{blue}Per-sample explainability dashboard generated from the Feature Subgraph Module. Panels: multi-lead ECG with time highlights (left), node-importance heatmap and edge-importance histogram (top/right), and prediction summary (bottom). Salient time windows appear coherently across leads, node importance is concentrated within a small subset of patches, and the edge distribution frequently exhibits a long tail, indicating sparse but critical relational evidence supporting the decision.}
  \label{fig:appB1}
  
\end{figure*}
}

